\documentclass[onefignum, onetabnum,onealgnum, onethmnum]{siamonline190516}

\usepackage{pifont}

\usepackage{lipsum}

\usepackage{endnotes}

\usepackage{titlesec}
\usepackage{titletoc}

\titleclass{\task}{straight}[\section]
\newcounter{task}
\renewcommand{\thetask}{\arabic{task}}
\titleformat{\task}[hang]
    {\normalfont\LARGE\bfseries}{Task \thetask:}{1em}{}

\titleformat*{\task}{\color{header1}\bfseries}

\titlecontents{task}
              [3.8em] 
              {}
              {\contentslabel{2.3em}}
              {\hspace*{-2.3em}}
              {\titlerule*[1pc]{.}\contentspage}

\titlespacing*{\section}{0ex}{1ex}{1ex}
\titlespacing*{\subsection}{0ex}{1ex}{1ex}
\titlespacing*{\paragraph}{0ex}{1ex}{1ex}
\titlespacing*{\subparagraph}{0pt}{1ex}{1ex}
\titlespacing*{\task}{0em}{1ex}{1ex}

\usepackage[sort&compress,comma,square,numbers]{natbib}

\usepackage{amsfonts,amsmath,amssymb}
\usepackage{bbm}
\usepackage{xspace}

\usepackage{multicol}
\usepackage{enumitem}

\usepackage{hhline}

\usepackage{comment}
\usepackage{todonotes}
\RequirePackage{ulem}

\usepackage{booktabs}
\usepackage{titlesec}
\usepackage{fancyhdr}
\usepackage{fullpage}

\RequirePackage{titlesec}
\RequirePackage[font={small},{singlelinecheck=false}]{caption}
\usepackage[T1]{fontenc}
\usepackage[scaled]{helvet}
\usepackage[scaled=1.10, med]{zlmtt}

\usepackage{algorithm}
\usepackage{algpseudocode}

\usepackage[super]{nth}
\usepackage{mdframed}
\mdfdefinestyle{MyFrame}{%
    outerlinewidth=6pt,
    roundcorner=20pt,
    innertopmargin=\baselineskip,
    innerbottommargin=\baselineskip,
    innerrightmargin=20pt,
    innerleftmargin=20pt 
    }

\providecommand{\sct}[1]{{\sc \texttt{#1}}}

\newcommand{\Mgc}{\sct{Mgc}}

\providecommand{\ve}[1]{\boldsymbol{#1}}
\providecommand{\ma}[1]{\boldsymbol{#1}}

\providecommand{\mc}[1]{\mathcal{#1}}
\providecommand{\mb}[1]{\boldsymbol{#1}}

\providecommand{\mt}[1]{\widetilde{#1}}

\providecommand{\mv}[1]{\vec{#1}}
\providecommand{\mh}[1]{\hat{#1}}

\providecommand{\mtb}[1]{\widetilde{\boldsymbol{#1}}}

\newcommand{\conv}{\rightarrow}

\newcommand{\mcE}{\mathcal{E}}

\newcommand{\defeq}{\coloneqq}

\newcommand{\PP}{\mathbb{P}}

\DeclareMathOperator{\R}{R} 
\DeclareMathOperator{\Y}{\mathbf{Y}}

\DeclareMathOperator{\X}{\mathbf{X}}

\newcommand{\bth}{\ve{\theta}}

\newcommand{\thetn}{\ve{\theta}}
\newcommand{\thet}{\thetn}

\usepackage{makecell}

\usepackage{listings}
\usepackage{xcolor}

\definecolor{codegreen}{rgb}{0,0.55,0}
\definecolor{codegray}{rgb}{0.5,0.5,0.5}
\definecolor{backcolor}{RGB}{245,245,245}
\definecolor{codepurple}{rgb}{.58,0,0.82}
\definecolor{JHBlue}{RGB}{0,45,114}

\lstdefinestyle{mystyle}{
    backgroundcolor=\color{backcolor},   
    commentstyle=\color{codegreen},
    keywordstyle=\color{JHBlue},
    numberstyle=\tiny\color{codegray},
    basicstyle=\ttfamily\footnotesize,
    breakatwhitespace=false,         
    breaklines=true,                 
    captionpos=b,                    
    keepspaces=true,                 
    numbers=left,                    
    numbersep=6pt,                  
    showspaces=false,                
    showstringspaces=false,
    showtabs=false,                  
    tabsize=2,
    frame=single
}

\newcommand{\mvlearn}{\texttt{mvlearn}}
\newcommand{\moduleskip}{0.1in}
\newcommand{\modulehanging}{.4in}

\usepackage{xcolor}

\title{mvlearn: Multiview Machine Learning in Python}
       
\author{%
    Ronan Perry$^1$,
    Gavin Mischler$^8$,
    Richard Guo$^2$,
    Theodore Lee$^1$,
    Alexander Chang$^1$,
    Arman Koul$^1$,
    Cameron Franz$^2$,
    Hugo Richard$^5$,
    Iain Carmichael$^6$,
    Pierre Ablin$^7$,
    Alexandre Gramfort$^5$, and
    Joshua T. Vogelstein$^{1,3,4}$
    \thanks{Corresponding author: 
    \href{mailto:jovo@jhu.edu}{jovo@jhu.edu}
        $^1$ Department of Biomedical Engineering
        Johns Hopkins University,
        $^2$ Department of Computer Science,
        Johns Hopkins University,
        $^3$ Center for Imaging Science,
        Institute for Computational Medicine,
        Kavli~Neuroscience~Discovery Institute,
        Johns Hopkins University, 
        $^4$ Progressive Learning
        $^5$ Universit\'e Paris-Saclay, Inria, Palaiseau, France
        $^6$ Department of Statistics, University of Washington, Seattle, WA 98195
        $^7$ CNRS and DMA, \'Ecole Normale Sup\'erieure, PSL University, Paris, France
        $^8$ Department of Electrical Engineering, Columbia University, New York, NY 10027
    } 
}

\begin{document}

\maketitle

\begin{abstract}
As data are generated more and more from multiple disparate sources, multiview data sets, where each sample has features in distinct views, have grown in recent years. However, no comprehensive package exists that enables non-specialists to use these methods easily. 
\mvlearn\, is a Python library which implements the leading multiview machine learning methods. Its simple API closely follows that of \texttt{scikit-learn} for increased ease-of-use. The package can be installed from Python Package Index (PyPI) and the \texttt{conda} package manager and is released under the MIT open-source license. The documentation, detailed examples, and all releases are available at \url{https://mvlearn.github.io/}.
\end{abstract}


\section{Introduction}
Multiview data (sometimes referred to as multi-modal, multi-table, or multi-block data), in which each sample is represented by multiple views of distinct features, are often seen in real-world data and related methods have grown in popularity. A view is defined as a partition of the complete set of feature variables \citep{xu2013survey}. Depending on the domain, these views may arise naturally from unique sources, or they may correspond to subsets of the same underlying feature space. For example, a doctor may have an MRI scan, a CT scan, and the answers to a clinical questionnaire for a diseased patient. However, classical methods for inference and analysis are often poorly suited to account for multiple views of the same sample, since they cannot properly account for complementing views that hold differing statistical properties \citep{zhao2017multi}. To deal with this, many multiview learning methods have been developed to take advantage of multiple data views and produce better results in various tasks \citep{sun2013survey, hardoon2004canonical, chao2017survey, yang2014automatic}.

Although multiview learning techniques are increasingly utilized in literature, no open-source Python package exists which implements an extensive variety of methods. The most relevant existing package, \texttt{multiview} \citep{kanaan2019multiview}, only includes 3 algorithms with an inconsistent API. \mvlearn\, fills this gap with a wide range of well-documented algorithms that address multiview learning in different areas, including clustering, semi-supervised classification, supervised classification, and joint subspace learning. Additionally, \mvlearn\, preprocessing tools can be used to generate multiple views from a single original data matrix, expanding the use-cases of multiview methods and potentially improving results over typical single-view methods with the same data \citep{sun2013survey}. Subsampled sets of features have notably led to successful ensembles of independent single-view algorithms \citep{ho1998random} but can also be taken advantage of jointly by multiview algorithms to reduce variance in unsupervised dimensionality reduction \citep{foster2008} and improve supervised task accuracy \citep{nigam2000analyzing}. The last column of Table \ref{Table 1} details which methods may be useful on single-view data after feature subsampling. \mvlearn\, has been tested on Linux, Mac, and PC platforms, and adheres to strong code quality principles. Continuous integration ensures compatibility with past versions, \texttt{PEP8} style guidelines keep the source code clean, and unit tests provide over 95\% code coverage at the time of release.


\begin{table}[!tb]
\begin{center}
  \resizebox{\linewidth}{!}{%
    \tabcolsep=0.11cm
    \begin{tabular}{ | c | c | c | c | }
      \hline
      \thead{\textbf{Module}} & \thead{\textbf{Algorithm (Reference)}} & \thead{\textbf{Maximum} \\ \textbf{Views} }& \thead{\textbf{Useful on} \\ \textbf{Constructed} \\ \textbf{Data from a} \\ \textbf{Single} \\ \textbf{Original View}} \\ [0.5ex]
      \hline\hline
      Decomposition & \thead{AJIVE \citep{feng2018angle}} & 2 & {\ding{55}}\\
     Decomposition & \thead{Group PCA/ICA \citep{calhoun2001method}} & $\geq$ 2 & {\ding{55}}\\
      Decomposition & \thead{Multiview ICA \citep{richard2020mvica}} & $\geq$ 2 & {\ding{55}}\\
      \hline
      Cluster & \thead{MV K-Means \citep{bickel2004multi}} & 2 & \ding{51} \\
      Cluster & \thead{MV Spherical K-Means \citep{bickel2004multi}} & 2 & \ding{51} \\
      Cluster & \thead{MV Spectral Clustering \citep{kumar2011cotrain}} & $\geq$ 2 & \ding{51} \\
      Cluster & \thead{Co-regularized MV Spectral Clustering \citep{kumar2011coreg}} & $\geq$ 2 & \ding{51} \\
      \hline
      Semi-supervised & \thead{Co-training Classifier \citep{blum1998combining}} & 2 & \ding{51} \\
      Semi-supervised & \thead{Co-training Regressor \citep{zhou2005semi}} & 2 & \ding{51} \\
      \hline
      Embed & \thead{CCA \citep{hotelling1992relations}} & 2 & {\ding{55}}  \\
      Embed & \thead{Multi CCA \citep{tenenhaus2011}} & $\geq$ 2 & {\ding{55}} \\
      Embed & \thead{Kernel Multi CCA \citep{hardoon2004canonical}} & $\geq$ 2 & {\ding{55}}  \\
      Embed & \thead{Deep CCA \citep{andrew2013deep}} & 2 & {\ding{55}}\\
      Embed & \thead{Generalized CCA \citep{afshin2012enhancing}} & $\geq$ 2 & {\ding{55}} \\
      Embed & \thead{MV Multi-dimensional Scaling (MVMDS) \citep{trendafilov2010stepwise}} & $\geq$ 2 & {\ding{55}} \\
      Embed & \thead{Omnibus Embed \citep{levin2017central}} & $\geq$ 2 & {\ding{55}} \\
      Embed & \thead{Split Autoencoder \citep{wang2015deep}} & 2 & {\ding{55}} \\
      \hline
    \end{tabular}}
    \caption{\textbf{Multiview (MV) algorithms offered in \mvlearn\, and their properties.}}
    \label{Table 1} 
  \end{center}
\end{table}

\section{API Design}
The API closely follows that of \texttt{scikit-learn} \citep{pedregosa2011scikit} to make the package accessible to those with even basic knowledge of machine learning in Python \citep{buitinck2013api}. The main object type in \mvlearn\, is the estimator object, which is modeled after \texttt{scikit-learn}'s estimator. \mvlearn\, changes the familiar method \texttt{fit(X, y)} into a multiview equivalent, \texttt{fit(Xs, y)}, where \texttt{Xs} is a list of data matrices, corresponding to a set of views with matched samples (i.e. the $i$\textsuperscript{th} row of each matrix represents the features of the same $i$\textsuperscript{th} sample across views). Note that \texttt{Xs} need not be a third-order tensor as each view need not have the same number of features. As in \texttt{scikit-learn}, classes which make a prediction implement \texttt{predict(Xs)}, or \texttt{fit\_predict(Xs, y)} if the algorithm requires them to be performed jointly, where the labels \texttt{y} are only used in supervised algorithms. Similarly, all classes which transform views, such as all the embedding methods, implement \texttt{transform(Xs, y)} or \texttt{fit\_transform(Xs)}.

\section{Library Overview}
\mvlearn\, includes a wide breadth of method categories and ensures that each offers enough depth so that users can select the algorithm that best suits their data. The package is organized into the modules listed below which includes the multiview algorithms in Table \ref{Table 1} as well as various utility and preprocessing functions. The modules' summaries describe their use and fundamental application.

\vspace{\moduleskip{}}
\hangindent=\modulehanging
\textbf{Decomposition:} \mvlearn\, implements the Angle-based Joint and Individual Variation Explained (AJIVE) algorithm \citep{feng2018angle}, an updated version of the JIVE algorithm \citep{lock2013joint}. This was originally developed to deal with genomic data and characterize similarities and differences between data sets. \mvlearn\, also implements multiview independent component analysis (ICA) methods~\citep{calhoun2001method, richard2020mvica}, originally developed for fMRI processing.

\vspace{\moduleskip{}}
\hangindent=\modulehanging
\textbf{Cluster:} \mvlearn\, contains multiple algorithms for multiview clustering, which can better take advantage of multiview data by using unsupervised adaptations of co-training. Even when the only apparent distinction between views is the data type of certain features, such as categorical and continuous variables, multiview clustering has been very successful \citep{chao2017survey}.

\vspace{\moduleskip{}}
\hangindent=\modulehanging
\textbf{Semi-supervised:}  Semi-supervised classification (which includes fully-supervised classification as a special case) is implemented with the co-training framework \citep{blum1998combining}, which uses information from complementary views of (potentially) partially labeled data to train a classification system. If desired, the user can specify nearly any type of classifier for each view, specifically any \texttt{scikit-learn}-compatible classifier which implements a \texttt{predict\_proba} method. Additionally, the package offers semi-supervised regression \citep{zhou2005semi} using the co-training framework.

\vspace{\moduleskip{}}
\hangindent=\modulehanging
\textbf{Embed:} \mvlearn\, offers an extensive suite of algorithms for learning latent space embeddings and joint representations of views. One category is canonical correlation analysis (CCA) methods, which learn transformations of two views such that the outputs are highly correlated. Many software libraries include basic CCA, but \mvlearn\, also implements several more general variants, including multiview CCA \citep{tenenhaus2011} for more than two views, Kernel multiview CCA \citep{hardoon2004canonical, bach2003, kuss2003}, Deep CCA \citep{andrew2013deep}, and Generalized CCA \citep{afshin2012enhancing} which is efficiently parallelizable to any number of views. Several other methods for dimensionality reduction and joint subspace learning are included as well, such as multiview multi-dimensional scaling \citep{trendafilov2010stepwise}, omnibus embedding \citep{levin2017central}, and a split autoencoder \citep{wang2015deep}.

\vspace{\moduleskip{}}
\hangindent=\modulehanging
\textbf{Compose:} Several functions for integrating single-view and multiview methods are implemented, facilitating operations such as preprocessing, merging, or creating multiview data sets. If the user only has a single view of data, view-generation algorithms in this module such as Gaussian random projections and random subspace projections allow multiview methods to still be applied and may improve upon results from single-view methods \citep{sun2013survey, nigam2000analyzing, ho1998random}.

\vspace{\moduleskip{}}
\hangindent=\modulehanging
\textbf{Data sets and Plotting:} A synthetic multiview data generator as well as dataloaders for the Multiple Features Data Set \citep{breukelen1998} in the UCI repository \citep{dua2019} and the genomics Nutrimouse data set \citep{nutrimouse2007} are included. Also, plotting tools extend \texttt{matplotlib} and \texttt{seaborn} to facilitate visualizing multiview data.

\section{Conclusion}
\mvlearn\, introduces an extensive collection of multiview learning tools, enabling anyone to readily access and apply such methods to their data. As an open-source package, \mvlearn\, welcomes contributors to add new desired functionality to further increase its applicability and appeal. As data are generated from more diverse sources and the use of machine learning extends to new fields, multiview learning techniques will be more useful to effectively extract information from real-world data sets. With these methods accessible to non-specialists, multiview learning algorithms will be able to improve results in academic and industry applications of machine learning.


\section*{Author Contribution}
Ronan Perry, Gavin Mischler — Conceptualization, API development, writing (initial draft), prototype codebase. Richard Guo, Theodore Lee, Alexander Chang, Arman Koul, Cameron Franz — Conceptualization, API development, prototype codebase. Hugo Richard, Pierre Ablin, Alexandre Gramfort —  API updates, major code revisions, writing (review and edits). Iain Carmichael — methodology, major code revisions, writing (review). Joshua T. Vogelstein — Conceptualization, supervision, funding acquisition, methodology, writing (review and edits).

\section{Acknowledgements}
This work is supported by the Defense Advanced
Research Projects Agency (DARPA) Lifelong Learning Machines program through contract FA8650-18-2-7834 and through funding from Microsoft Research. We thank all the contributors for assisting with writing \mvlearn. We thank the NeuroData Design class and the NeuroData lab at Johns Hopkins University for support and guidance.

\bibliographystyle{unsrtnat}
\bibliography{references}

\end{document}